\pdfoutput=1

\documentclass[11pt]{article}

\usepackage{acl}

\usepackage{times}
\usepackage{latexsym}
\usepackage{tabularx}
\usepackage{soul}

\usepackage[T1]{fontenc}

\usepackage[utf8]{inputenc}

\usepackage{microtype}
\usepackage{algorithmicx}

\usepackage{algpseudocodex}
\usepackage{algorithm}
\usepackage{float}

\usepackage{array}
\usepackage{amsthm}
\usepackage{amssymb}
\usepackage{amsmath}
\usepackage{bbold}

\DeclareMathOperator*{\argmax}{arg\,max}

\theoremstyle{plain}

\theoremstyle{definition}
\newtheorem{defn}{Definition}
\newtheorem{example}{Example}

\theoremstyle{remark}

\newcommand{\Crond}{\mathcal C}

\newcommand{\Nrond}{\mathcal N}
\newcommand{\Orond}{\mathcal O}

\newcommand{\Urond}{\mathcal U}

\newcommand{\vertiii}[1]{{\left\vert\kern-0.25ex\left\vert\kern-0.25ex\left\vert #1
\right\vert\kern-0.25ex\right\vert\kern-0.25ex\right\vert}}

\newcommand{\set}[1]{\left\{ #1\right\}}

\renewcommand{\leq}{\leqslant}

\usepackage{xcolor}
\usepackage{tcolorbox}
\usepackage{subcaption}
\usepackage{multirow}
\usepackage{booktabs}
\usepackage{enumitem}

\usepackage{pgfplotstable}

\usepackage{xspace}
\usepackage{inconsolata}
\usepackage{algpseudocode}
\newcommand{\sys}{GLIMPSE\xspace}

\usepackage{hhline}

\newtcolorbox{coloredbox}[3][]
{
  colframe = #2!25,
  colback  = #2!10,
  coltitle = #2!20!black,  
  title    = {#3},
  #1,
}

\newcommand{\hlc}[2]{{%
    \definecolor{foo}{rgb}{#1}
    \sethlcolor{foo}\hl{#2}}%
}

\definecolor{darkgray}{gray}{0.00}

\title{\sys: Pragmatically Informative Multi-Document Summarization of Scholarly Reviews}

\pgfplotstableread{image.dat}{\pixeldata}

\author{ {$^{\ast}$\bf Maxime \textsc{Darrin}$^{1, 2, 3, 4}$} \quad {$^\ast$\bf Ines \textsc{Arous}$^{ 2, 3}$} \\ {\bf Pablo \textsc{Piantanida}$^{1, 2, 4, 5}$} \quad {\bf Jackie Chi Kit \textsc{Cheung}$^{2, 3, 6}$} \\  $^{1}$International Laboratory on Learning Systems, $^{2}$MILA - Quebec AI Institute \\ $^{3}$McGill University $^{4}$Université Paris-Saclay \\ $^{5}$CNRS, CentraleSupélec, $^6$Canada CIFAR AI Chair \\ { \url{maxime.darrin@mila.quebec} \quad \url{ines.arous@mila.quebec} } \\ { \url{pablo.piantanida@mila.quebec} \quad \url{jackie.cheung@mcgill.ca}} }

\begin{document}

\maketitle
\begin{abstract}
Scientific peer review is essential for the quality of academic publications. However, the increasing number of paper submissions to conferences has strained the reviewing process. This surge poses a burden on area chairs who have to carefully read an ever-growing volume of reviews and discern each reviewer's main arguments as part of their decision process. In this paper, we introduce \sys, a summarization method designed to offer a concise yet comprehensive overview of scholarly reviews. Unlike traditional consensus-based methods, \sys extracts both common and unique opinions from the reviews. We introduce novel uniqueness scores based on the Rational Speech Act framework to identify relevant sentences in the reviews. Our method aims to provide a pragmatic glimpse into all reviews, offering a balanced perspective on their opinions. Our experimental results with both automatic metrics and human evaluation show that \sys generates more discriminative summaries than baseline methods in terms of human evaluation while achieving comparable performance with these methods in terms of automatic metrics.
\end{abstract}

\def\thefootnote{*}\footnotetext{Equal contribution.}\def\thefootnote{\arabic{footnote}}

\section{Introduction}
Peer review is the standard process for evaluating researchers' work submitted to conferences or academic journals across all fields. Its primary function is to maintain quality standards for academic publications and provide authors with constructive feedback on their work. Its effectiveness is currently being challenged by a significant surge in the number of submissions. Conferences in computer science, such as the International Conference on Learning Representations (ICLR) and the Association for Computational Linguistics (ACL), among others, regularly receive thousands of submissions. For instance, the number of submissions to ICLR and ACL has increased fivefold since 2017, reaching 3407 and 3378 submissions, respectively, in 2022.

The reviewing process is inherently a tool for scientific communication with at least two target audiences: the authors of a paper and the area chair. The former should get feedback on their work, whereas the latter has to synthesize the salient points from all the reviews as part of their decision process. The area chair has to extract from the reviews the overall sentiment about the paper while gathering the common ideas and unique arguments raised by the reviewers. An efficient highlighting mechanism could help area chairs in their decision-making process, thus reducing their workload.

Several methods have been developed to generate summaries from reviews in various domains~\cite{brazinskas_unsupervised_2020,chu_meansum_2019}. Many of these methods identify salient segments in reviews based on the discussed topics and the sentiment polarity~\cite{zhao_weakly-supervised_2020,li_aspect-aware_2023,amplayo_aspect-controllable_2021}, or using centrality-based metrics~\cite{ge_finesum_2023,liang_improving_2021}.
However, these techniques fall short in the peer review domain. Indeed, they are designed to generate a \emph{consensus} opinion summary, reflecting common opinions without identifying divergent and unique ones. This poses a challenge for the peer review domain since area chairs are concerned with both common and divergent opinions among reviewers. %

In this paper, we recognize one of the underlying communication goals in the reviewing/meta-reviewing process: to convey the review's main points to the area chair concisely. The distillation of such salient information in a concise message has been a long-standing focus of study within the pragmatics domain. One of the most influential probabilistic approaches
to pragmatics is the \textbf{Rational Speech Act} (RSA). It formulates the communication problem akin to a "reference game", with the goal of associating each item in a set with the most informative yet concise utterance that distinguishes it from others.

We take inspiration from this reference-game scenario to introduce the discriminative summarization task. The goal is to generate a summary for each review that highlights commonalities, differences, and unique perspectives, thereby distinguishing one review from others of the same submission. This framework closely aligns with the challenge of summarizing academic reviews, distilling key discriminative insights from lengthy reviews, thus contextualizing each review in relation to others. To this end, we map the  discriminative summarization task to a reference game and propose \textbf{\sys, a novel pragmatically informative summarization method for scholarly reviews.} At the technical level, we leverage the RSA model, a framework for pragmatic modeling rooted in Bayesian inference that solves the reference game setting~\cite{frank_predicting_2012}. We define two novel RSA-based scores that measure the \emph{informativeness} and the \emph{uniqueness} of opinions in scholarly reviews. We use these scores to rank utterances describing a review and aggregate them to compose a ``glimpse'' of all reviews.

We conducted extensive experiments on a real-world peer review dataset from the ICLR conference collected over a four-year time period. We compare \sys performance to state-of-the-art methods in multi-document summarization. We design extractive and abstractive variants of our framework and shed light on their properties. Our results show that \sys generates informative and concise summaries. To the best of our knowledge, we are the first to cast the multi-document summarization problem as a reference game and adopt RSA to identify common and divergent opinions in reviews. 

\textbf{Contributions:} Overall, we make the following key contributions:

\begin{enumerate}[noitemsep,topsep=0pt,parsep=0pt,partopsep=0pt]
\item We propose a new setting for multi-document summarization: discriminative summarization and cast it as a reference game problem.
    \item We propose new \emph{informativeness} and \emph{uniqueness} scores for multi-document summarization based on the RSA model of human communication\footnote{\url{https://github.com/icannos/glimpse-mds}}.
    \item We conduct empirical evaluation to demonstrate that we extract more discriminative summaries than consensus-based summarizers in both automatic and human evaluation.
\end{enumerate}

\begin{figure*}
    \centering

    \resizebox{0.8\textwidth}{!}{
\begin{tabularx}{\textwidth}{X}
    \hlc{0.88,0.9,0.93}{Solid theoretical results are provided to confirm the doubly robustness of the treatment estimator
and outcome estimator.} \hlc{0.72,0.78,0.89}{This paper is well-written, however, I still have some concerns about the
contributions.} 
[...]
\hlc{0.91,0.79,0.75}{Why the VCnet designed to be dependent of treatment information?} \hlc{0.81,0.84,0.91}{The dependence is not theoretically discussed in this paper.} \hlc{0.75,0.8,0.9}{The experimental results are not convincing
for me.} \hlc{0.79,0.83,0.91}{Only two baselines are compared with the proposed method.} %
[...]
\hlc{0.93,0.85,0.83}{But the design of
VCnet needs the numerical results to confirm its effectiveness for ADRF.}  \\ \hline \hline
    \hlc{0.91,0.78,0.74}{
This paper is to develop a varying coefficient neural network to estimate average dose-response curve (ADRF).} \hlc{0.91,0.8,0.77}{Although this paper has several interesting results, the paper is full of many typos and small errors.} \hlc{0.92,0.81,0.77}{The current paper needs substantial improvement.} \hlc{0.91,0.77,0.72}{The introduction section is not well written since the logic does not flow very smooth..} \hlc{0.91,0.8,0.76}{In the proof of all theorems, there are some obvious mistakes inside.}  \\ \hline \hline
    \hlc{0.95,0.94,0.94}{
This problem is well-motivated - estimating dose-response is a challenging and practically important
problem.} \hlc{0.82,0.85,0.92}{The paper is well written.} \hlc{0.93,0.87,0.85}{It explained complex ideas in the semi-
parametric literature clearly.} \hlc{0.92,0.93,0.94}{The comparison against existing works is clear.} \hlc{0.88,0.9,0.93}{The theory, as far as
I can tell, is solid.} 
\hlc{0.92,0.83,0.8}{It improves the existing results in targeted regularization and can be adapted to analyze one step TMLE.} 
\end{tabularx}}
    \caption{
    Illustration of our proposed RSA-based scores applied to real-world scholarly reviews. We consider each sentence in a review as a candidate summary.
    The most common opinions in the reviews are highlighted in blue whereas the unique ones are highlighted in red using our RSA-based scores.}
    \label{fig:dummy_review_example}
\end{figure*}

\section{Related Work}

\noindent\textbf{Unsupervised Opinion Summarization.} 
Our work is related to unsupervised opinion summarization, where several methods have been developed to summarize reviews of products or hotels.
These methods can be divided into two categories: abstractive and extractive. Abstractive summarization aims to generate a coherent summary that reflects salient opinions in input reviews. For instance, MeanSum~\cite{chu_meansum_2019} relies on an auto-encoder architecture, in which an aggregated representation of the input reviews is fed to a decoder that generates review-like summaries. 
Another approach consists of using a hierarchical variational autoencoder model~\cite{brazinskas_unsupervised_2020} to generate summaries that represent dominant opinions within reviews. Other methods rely on modeling fine-grained information within reviews. This includes disentangling aspects and sentiments~\cite{amplayo_unsupervised_2021,wang_transsum_2021,suhara_opiniondigest_2020,ke_consistsum_2022}, topics~\cite{xu_coarse--fine_2020} and contrastive opinions~\cite{iso_comparative_2022,carenini_extractive_2008}. 
The absence of attributability in these methods presents a challenge in the peer review domain due to the need for transparency and context. In the peer review process, area chairs might seek to trace the source of the synthesized opinions to assess their validity and relevance. \textbf{Our method addresses the challenge of attributability} by associating a summary with each review, ensuring transparency and allowing area chairs to trace and understand the synthesized opinions in context.

In contrast to abstractive summarization, which entails generating new text, extractive summarization aims to extract significant phrases directly from the input reviews. A fundamental method in extractive summarization involves clustering review segments and iteratively extracting the central elements of these segments to form a summary. Various techniques depend on a model to acquire representations for review sentences. These representations are then utilized by an inference algorithm, often in an encoder-decoder architecture, to choose sentences for summarization~\cite{li_aspect-aware_2023,amplayo_aspect-controllable_2021,angelidis_extractive_2021,gu_memsum_2022, angelidis_summarizing_2018}. Current extractive methods concentrate on amalgamating common opinions found across reviews. However, we contend that such common opinions hold less significance in scholarly contexts. Instead, we advocate for a pragmatic approach that situates each review relative to others, emphasizing unique, common, and divergent ideas~\cite{mani_summarizing_1999, wan_summarizing_2011}.

\noindent\textbf{Summarization in Scientific Peer Review}
The task most closely related to ours in the scholarly domain is the meta-review generation task. 
Various strategies have been developed for this task. For instance, MetaGen~\cite{bhatia_metagen_2020} generates an extractive summary draft of reviews, then uses a fine-tuned model for decision prediction, and generates an abstractive meta-review based on both the draft and the predicted decision. Similarly,~\cite{li_towards_2023} leverage the hierarchical relationships within reviews and metadata, such as reviewers' confidence and rating, to generate a meta-review along with the acceptance decision. Recently, ~\cite{zeng_scientific_2023} proposed to prompt a large language model in a guided and iterative manner to generate a meta-review. Our task is different because our goal is to support area chairs by generating a summary of reviewers' opinions and highlighting areas of divergence. Predicting the meta-reviewer's decision and generating a meta-review based on the predicted decision is out of the scope of our work as we believe it is not desirable to do so. The metareview evaluation should ultimately remain in the hands of a human expert as it involves scientific expertise and reasoning about the evaluated paper.

\noindent\textbf{Rational Speech Act (RSA) framework.} The Rational Speech Act theory~\cite{frank_predicting_2012} is the most influential probabilistic approach to pragmatics~\cite{qing_variations_2015,degen_rational_2023}. It formulates communication between a pragmatic speaker and a listener as probabilistic reasoning, where the speaker’s goal is to choose the utterance that is both short and informative with respect to the speaker’s intended referent~\cite{degen_rational_2023}. As a result, the RSA framework aims to effectively identify the most informative utterance from multiple potential options in a given context. It has been applied to various tasks, including image captioning~\cite{ou_pragmatic_2023,cohn-gordon_pragmatically_2018}, translation~\cite{cohn-gordon_lost_2019}, dialogue~\cite{kim_will_2020,kim_perspective-taking_2021}, and text generation~\cite{shen_pragmatically_2019}, among others. However, there is little research on applying the RSA framework to multi-document summarization. In our study, we tailor the RSA framework for multi-document summarization tasks, aiming to produce summaries that capture various perspectives while minimizing redundancy.

\section{Discriminative Multi-Document Summarization (D-MDS)}

Most approaches to multi-document summarization prioritize generating consensus-based summaries by emphasizing redundant opinions across reviews. However, in domains like peer review, where including unique and divergent opinions is crucial for a comprehensive summary, the conventional focus on common opinions becomes less relevant. This shift in emphasis is particularly pertinent in scholarly review summarization, where the objective is to convey the various opinions and distinctive viewpoints expressed by reviewers. With this objective in mind, we introduce the discriminative multi-document summarization task, inspired by the reference game setting. \textbf{Our aim is to furnish the meta-reviewer with a summary for each review, enabling them to swiftly identify the source review based on the summary content.}

Formally, we define the discriminative multi-document summarization problem as follows:  
\begin{defn}[The discriminative multi-document summarization problem]
Let $\mathcal{N}=\set{d_1, \cdots, d_N}$ be a set of documents. For each document $d_i$, we suppose we have $K$ candidate summaries and we form $\mathcal{K} = \set{s_{i,j}}_{1 \leq i \leq N \\, 1 \leq j\leq K}$ the set of all candidates.
The goal is to select the most informative summary from that set for each document $d_i \in \Nrond$
\end{defn}

These candidates can be generated using various summarization methods (see \autoref{sec:candidates_generation}).

\section{Problem Formulation and Pragmatic Summarization}

In this section, we formulate the discriminative multi-document summarization problem as a reference game. We then apply the Rational Speech Act (RSA) framework, a probabilistic approach to pragmatics that tackles reference games, to address this summarization problem.

\subsection{D-MDS Problem as a Reference Game}
In a reference game setting~\cite{frank_predicting_2012}, a speaker and a listener are given a set of objects. The speaker provides a description of a target among the set of objects, and a listener aims to select the correct target given the speaker's description. The speaker's goal is to describe the target using one of its properties in an informative yet concise manner. Similarly, in discriminative multi-document summarization, our goal is to develop a summarizer that provides a concise yet informative description of each review within a set of reviews. Given this similarity, we formally map the D-MDS problem to a reference game setting as follows.
\begin{defn}[D-MDS as a reference game]
    \label{def:reference_game}
    Let $\Orond \triangleq  \set{d_1, \cdots, d_N}$ be the set of documents and $\Crond \triangleq  \set{s_{i,j}}_{1 \leq i \leq N \\, 1 \leq j\leq K}$ the set of candidate summaries.
   Let $M : \Orond \times \Crond \rightarrow [0, 1]$  be a truth matrix that indicates the likelihood of a candidate summary $s$ being a summary of a document $d$.
    In standard reference games, the truth-matrix is boolean as the properties for an object are either true or false. In our setting, we approximate the truth-matrix using a pre-trained language model on summarization $\operatorname{LM}$ to score the likelihood of each candidate summary $s$ to be associated with a document $d$: $M(t, s) \approx \operatorname{LM}(s | d)$.
\end{defn}

\subsection{The Pragmatic Summarizer}
\label{sec:rsa_reranking_method}

One popular framework to efficiently tackle reference games is the Rational Speech Act (RSA) model of human communication~\citet{frank_predicting_2012}, which models optimal communication between pragmatic agents and provides a formal framework to build pragmatic speakers and listeners. Therefore, we propose two novel RSA-based scores to select informative and unique opinions for review summarization, which we discuss in detail in \autoref{sec:rank}. In what follows, we present how we adapt RSA to our summarization setting. 

The RSA framework posits that both the speaker and listener maximize the utility of their communication and they both assume the other is rational. They iteratively adapt to each other to reach a common understanding of the context.
Formally, we define the utility of communicating to a listener $L$, the summary $s$ for a document $d$ as:
\begin{align}
    V_L(s, d) = \log L(d | s) - \operatorname{Cost}(d),
    \label{eq:utility_base}
\end{align}
where $L(d | s)$ denotes the conditional probability of guessing the object $d$ --- the document/review in our case --- upon receiving $s$ --- a short summary--- according to the listener $L$, $\operatorname{Cost}(s)$ is the cost of transmitting the summary $s$. In most cases, the cost is assumed to be $0$, and we measure the informativeness, which is defined with the conditional probability $L(d | s)$. 

Pragmatic reasoning is modeled as an iterative process that starts from a \emph{literal} listener, \textit{i.e.} a listener who has no assumption about the speaker. We denote it with $L_0(d | s)$ and define it using a pre-trained language model $\operatorname{LM}$:
\begin{defn}[Literal listener]
    \label{def:literal_listener}
    \begin{align}
    \label{eq:literal_listener}
        L_0(d | s) = \frac{\operatorname{LM}(s | d)}{ \sum_{d \in \Orond} \operatorname{LM}(s | d) }. 
    \end{align}
\end{defn}

The pragmatic speaker adapts to a listener under the cost constraint by maximizing the communication utility~\cite{zaslavsky_ratedistortion_2021}. We define it as follows:
\begin{defn}[Pragmatic speaker]
\label{def:speaker}
 For a given target document $d$, the speaker's distribution over the summaries is defined as the softmax of the utility scores with a listener $L_{t-1}$, where $L_{0}$ is the literal listener given in~\autoref{eq:literal_listener}.
\begin{align}
\label{eq:pragmatic_speaker}
        S_t(s | d) &= \frac{\exp(V_{L_{t-1}}(d, s))}{\sum_{s'} \exp(V_{L_{t-1}}(d, s')}.
\end{align}
\end{defn}

The pragmatic listener evaluates the summary provided by the pragmatic speaker and identifies the document that is most likely to be its source. 
\begin{defn}[Pragmatic listener]
\label{def:pragmatic_listener} 
 For a given target summary $s$, the pragmatic listener's distribution over the documents is defined using the pragmatic speaker $S_t$ from~\autoref{eq:pragmatic_speaker} as follows:
\begin{equation}
\label{eq:pragmatic_listener} 
        L_t(d | s) = \frac{S_t(s | d)}{\sum_{d'} S_t(s | d')}.
    \end{equation}
\end{defn}

We iterate between the pragmatic speaker (\autoref{eq:pragmatic_speaker}) and the listener (\autoref{eq:pragmatic_listener}), recursively adapting the listener to the speaker for an empirically defined number $T$ of iterations. 

\section{\sys Framework}

Our framework comprises three steps: 1) generating candidates through either abstractive or extractive summarization techniques, 2) selecting candidates using RSA-based scores defined in \autoref{sec:rank}, and 3) composing a summary using the selected candidates and a template.

\subsection{Candidate Generation}
\label{sec:candidates_generation}
The candidate summaries can be extracted sentences from the reviews, abstractive summaries sampled from a summarization model conditioned to review, or any other generated summary. We consider both the extractive setting, where the candidates are extracted sentences from the reviews, and the abstractive setting, where a language model generates the candidate summaries. We specifically focus on the extractive setting, as we want to ensure attribution of the summary to the source review.

\subsection{Informativeness and Uniqueness Scores}
\label{sec:rank}
We propose two RSA-based scores derived from the RSA speaker and listener in our setting \autoref{sec:rsa_reranking_method}: a \emph{pragmatic-speaker-based score} and a \emph{uniqueness score}. The \emph{pragmatic-speaker-based score} identifies the most discriminative utterance to refer to a source document. The \emph{uniqueness score} measures the extent to which a candidate summary represents a common --- or unique --- idea in the source documents.

\noindent\textbf{Pragmatic-speaker-based score.} We directly score the summaries for a document according to the pragmatic speaker's distribution. In our case, this process assesses the informativeness of a summary within the set of candidates.  Hence, we define the \emph{pragmatic-speaker-based score}  to be the argmax of the speaker: \autoref{eq:pragmatic_speaker}:
\begin{align}
    \operatorname{RSA-Speaker}(d) \triangleq  \argmax_{s\in \Crond} S_t(s | d).
    \label{eq:speaker_score}
\end{align}
In our experiments, we refer to the summarizer leveraging this score as \sys-Speaker.

\noindent\textbf{Uniqueness score.}
In our context, the RSA listener is designed to pragmatically infer a document given a summary. We propose quantifying the listener's uncertainty regarding a particular summary given the following intuition: {if the listener is uncertain, then the summary could apply to many documents; conversely, if the listener is confident, the summary can only be associated with a single document.}
\begin{example}
    Consider the following reviews.
    \begin{itemize}
        \item Review 1: \textit{This paper is well-written. However, the theoretical part lacks clarification.}
        \item Review 2: \textit{This paper is well-written. I believe it should be accepted.}
    \end{itemize}

    Given the sentence "\textit{The paper is well-written,}" a listener cannot accurately deduce the source review. However, when provided with the sentence "\textit{I believe it should be accepted,}" the listener can easily identify that the source review is Review 2.
\end{example}

We propose to measure this uncertainty by comparing the listener's probability distribution defined in \autoref{eq:pragmatic_listener} with the uniform distribution $\Urond$ over the documents:
\begin{equation}
    \operatorname{Unique} \triangleq  D_{\text{KL}}(L(\cdot |\ s) \| \Urond).
    \label{eq:disag_score}
\end{equation}
High values of uniqueness score indicate the uniqueness of the candidate summary or its divergence from other candidates while lower values indicate that the candidate summary is common to multiple source documents. 
Intuitively, it measures how far the listener distribution conditioned to a summary is far from the uniform distribution. The {\sys} variant using this score is referred to as {\sys}-Unique.

These two scores can be used to identify unique or common opinions across the different documents, as shown in~\autoref{fig:dummy_review_example}, or to select the most informative summary among a set of candidates. While, in this work, we use the scores to compose overall summaries, they can be used as standalone tools for content highlighting.

\subsection{{\sys} for Standard Multi-Document Summarization (MDS)}
\label{sec:MDS}

The standard Multi-Document Summarization (MDS) task aims to generate a single summary for multiple documents. Generally, this summary is built to reflect the consensus among the documents. In order to provide a comparison with previous work, we generate summaries by concatenating the three most common with the three most unique candidate summaries identified through our RSA-based scores defined in \autoref{eq:speaker_score} and \autoref{eq:disag_score}. The most unique candidates are selected either using the RSA speaker score  (\autoref{eq:speaker_score}) or using the uniqueness score (\autoref{eq:disag_score}). Summaries utilizing the former are referred to as \sys-Speaker, while those utilizing the latter are referred to as \sys-Unique. This method ensures a balanced representation of both common and unique opinions.

\begin{example}
    Given the reviews in \autoref{fig:dummy_review_example}, we can compose two summaries using our simple template:
    \begin{itemize}
        \item \textbf{\sys-Speaker}: This paper is well-written, however, I still have some concerns about the contributions. The experimental results are not convincing I really enjoyed reading it. The introduction section is not well written since the logic does not flow very smooth. Why the VCnet designed to be dependent of treatment information? 
        \item \textbf{\sys-Unique}: This paper is well-written, however, I still have some concerns about the contributions. The experimental results are not convincing I really enjoyed reading it.
        The introduction section is not well written since the logic does not flow very smooth. 
        This paper is to develop a varying coefficient neural network to estimate average dose-response curve (ADRF). Why the VCnet designed to be dependent of treatment information?
    \end{itemize}
\end{example}

\section{Experimental Setup}

In this section, we outline the experimental setup used to evaluate the performance of \sys. We present the datasets, baselines, evaluation metrics, and implementation details.

\noindent\textbf{Datasets.} We collect data from the ICLR conference, which provides open access to reviews and meta-reviews for all submissions through OpenReview\footnote{\href{https://openreview.net/}{https://openreview.net/}}. Our dataset contains 28062 reviews for 8428 submissions to the ICLR conference from 2017 until 2021.

\noindent\textbf{Evaluation.} The task of collecting reference summaries of scholarly reviews is challenging due to their length and domain specificity. 
We identify an alternative method to collect scholarly summaries by considering meta-reviews. Meta-reviews convey mainly the decision of the meta-reviewer and, depending on their style, may summarize the reviews. 
We only use these “summary-like” meta-reviews instead of those that solely convey area chairs’ decisions. 
We apply heuristic rules to extract “summary-like” meta-reviews and perform manual verification. We set criteria for identifying them, including length, reference to at least one reviewer, and semantic similarity measured by cosine similarity with all reviews. 
We obtained 226 summary-like meta-reviews that we use for evaluation.

\subsection{Evaluation Metrics} 
\noindent\textbf{Evaluation against gold standards.} For comprehensiveness and following common practices, we include ROUGE evaluations of the summaries against a gold standard. In our case, we assume that the area chair's motivations for their decision provide a reasonable comparison. However, this evaluation may be limited, because ROUGE scores have a low correlation with human judgments according to various studies~\cite{kryscinski_evaluating_2020, kocmi_ship_2021}.

\noindent\textbf{Discrimininativeness.} Following \cite{ou_pragmatic_2023, cohn-gordon_pragmatically_2018}, we measure the discriminativeness of a summary, i.e., whether an evaluative listener can identify the source review based on the summary content. We construct our evaluative listener by comparing the summary and the reviews using cosine
similarity of their paraphrase embeddings~\cite{reimers_sentence-bert_2019}.

\noindent\textbf{Learned metrics.} We evaluated the generated summaries using the SEAHORSE metrics trained on human judgment~\cite{clark_seahorse_2023}. They assess the summaries along six axes: coverage, attribution, comprehensibility, grammar, repetition and conciseness. Coverage and attribution, respectively, measure if the main ideas of the source text are present in the summary and verify the accuracy of information attribution. Comprehensibility and grammar assess the summary's overall fluency and grammatical correctness, while repetition and conciseness evaluate the absence of redundant utterances and the summary conciseness. In the multi-document summarization setting, we evaluate the summary generated as discussed in~\autoref{sec:MDS}. High coverage across all reviews suggests that the summary effectively captures the main ideas of all the reviews.

\subsection{Comparison Methods}
We compare our framework with baseline summarization methods: 1) Latent Semantic Analysis (LSA)~\cite{steinberger_using_2004} extracts sentences from a document based on latent topics. 2) LexRank~\cite{erkan_lexrank_2004}, is a graph algorithm that uses TF-IDF to calculate weights for sentence segments and selects segments near the center as the summary\footnote{LSA and LexRank are implemented using \url{github.com/miso-belica/sumy}}.  3) Quantized Transformer space (QT) proposes a clustering interpretation of the quantized space for extractive summarization~\cite{angelidis_extractive_2021}. 4) PlanSum~\cite{amplayo_informative_2021} is an abstractive summarization method that incorporates content planning in a summarization model. 5) Convex Aggregation for Opinion Summarization (COOP)~\cite{iso_convex_2021} is an abstractive summarization method that leverages embedding representation to condition summary generation. 6) Llama 7b Instruct\footnote{\url{https://huggingface.co/togethercomputer/Llama-2-7B-32K-Instruct}}, is a Large Language Model that generates abstractive summaries via prompting. In addition, we report the performance of the random selection of sentences from the documents. We also compare our method with generative models in terms of discriminativeness by evaluating the perplexity of candidate summaries conditioned on the source text.

\noindent\textbf{Candidate summary generation.} We generate candidate summaries by extracting sentences from reviews in the extractive setting. For the abstractive setting, we use pre-trained language models, such as PEGASUS~\cite{zhang_pegasus_2020} or BART summarizers~\cite{lewis_bart_2020}, to generate summaries for each review.

\section{Experimental Results}

We present the results for both the discriminative and the standard multi-document summarization tasks. We compare our method with the baselines and evaluate the quality of the generated summaries using automatic metrics and a human evaluation.

\begin{figure}
\begin{subfigure}{\linewidth}
            \includegraphics[width=\linewidth]{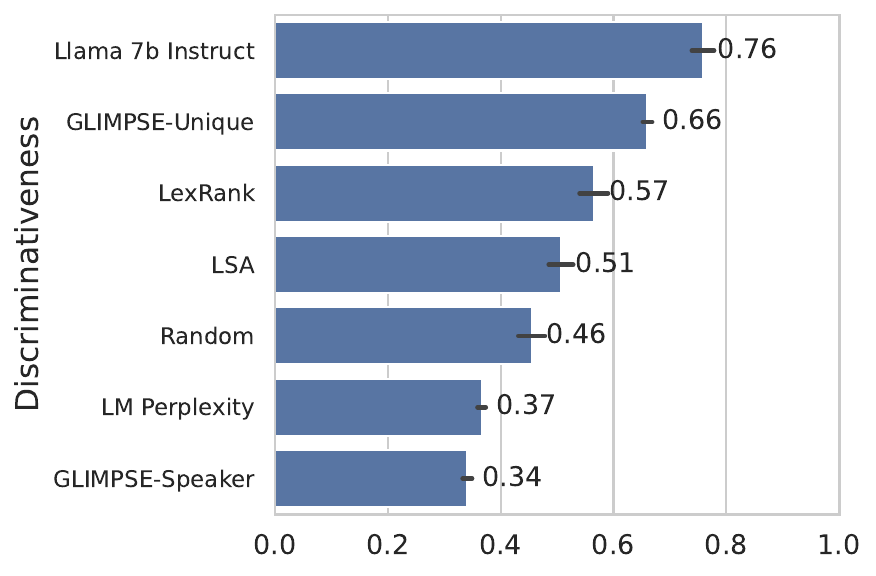}
            \caption{Discriminativeness.}
\end{subfigure}
    \begin{subfigure}{\linewidth}
            \includegraphics[width=\linewidth]{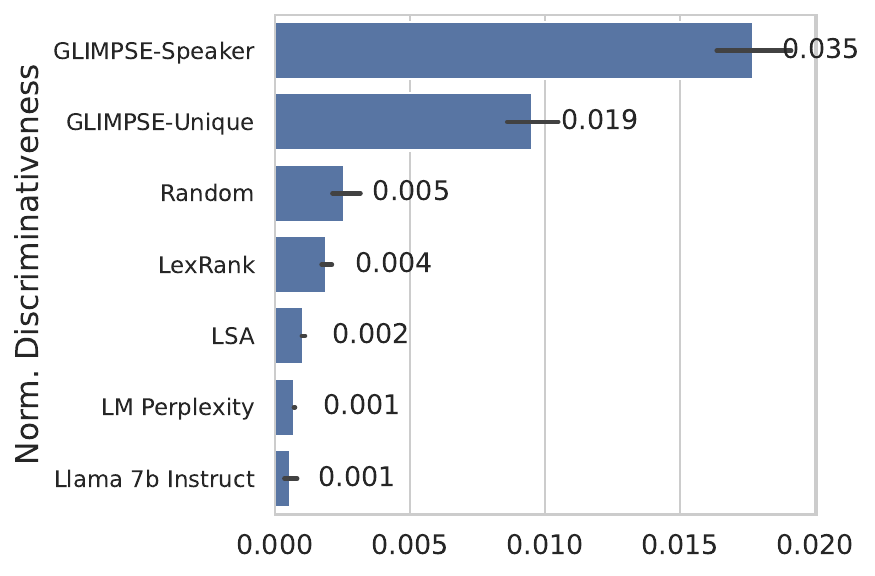}
    \caption{Discriminativeness per character.}
    \label{fig:baselines_informativeness_per_char}
    \end{subfigure}
    \caption{Discriminativeness for all the baselines and our methods in extractive mode ({\sys}-Unique (Extr.), {\sys}-Speaker (Extr.)), and a strong abstractive method (Llama 7b Instruct).}
    \label{fig:baselines_informativeness}
\end{figure}

\subsection{Discriminative Summarization}
\label{sec:discriminative_summ}

In this setting, our goal is to evaluate the \textbf{discriminativeness} of the generated summaries by gauging a listener's ability to identify the source review from the generated summary. 

\noindent\textbf{Discriminativeness evaluation.} In \autoref{fig:baselines_informativeness}, we report the discriminativeness scores for extractive methods alongside Llama 7b Instruct, a strong abstractive summarization baseline. We find that Llama 7b Instruct achieves the highest performance compared with other methods ($76\%$ discriminativeness); followed by {\sys}-Unique in the extractive setting ($66\%$ discriminativeness). These results suggest that our method achieves high discriminativeness while being more cost-effective compared to recent large language models in terms of memory footprint and runtime. Surprisingly, we find that selecting candidates using the perplexity of common abstractive summarizers, such as BART (LM Perplexity), yields results worse than a Random selection ($37\%$ vs $46\%$). Contrary to our expectations, {\sys}-Speaker has the lowest performance compared with baseline methods in terms of discriminativeness. We hypothesize that this result can be attributed to a phenomenon known as codebooking, where the RSA speaker overly tailors its output to the RSA listener, disregarding the semantic load of the summary~\cite{arumugam_rate-distortion_2022, wang_calibrate_2021}.

\noindent\textbf{Conciseness and information density.} When evaluating properties of the generated summaries, we observed a significant variation in the length of the generated summaries ranging from $76$ on average for \sys-Speaker to around $2000$ for Llama. We account for these differences by evaluating the "Discriminativeness per character" in  \autoref{fig:baselines_informativeness_per_char}. Interestingly, we find that the advantage of Llama vanishes completely, while \sys-Speaker and \sys-Unique notably outperform all baseline methods. This result suggests that {\sys} methods produce very succinct yet very discriminative summaries. This is key for the task at hand since the goal is to alleviate the strain on area chairs.

\begin{figure}[t]
    \centering    
    \includegraphics[width=0.45\textwidth]{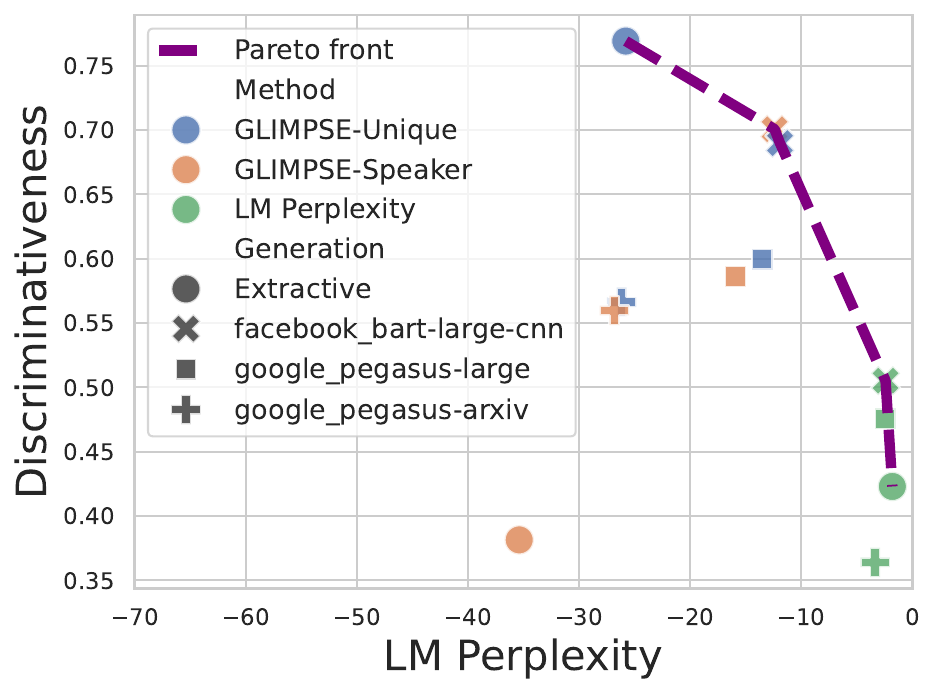}
    \caption{Trade-off between discriminativeness of the generated summaries and their fluency (measured as the log-likehood of the summaries under the generative model). The Pareto frontier shows the best trade-off between the two metrics.}
    \label{fig:trade_off_info_fluency}
\end{figure}
\begin{table*}

\resizebox{\textwidth}{!}{\begin{tabular}{lllrrrrrrrrrrrrrrrr}
\toprule
 &  &  & \multicolumn{2}{r}{Rouge-1} & \multicolumn{2}{r}{Rouge-2} & \multicolumn{2}{r}{Rouge-L} & \multicolumn{2}{r}{Cov.} & \multicolumn{2}{r}{Attr.} & \multicolumn{2}{r}{Gram.} & \multicolumn{2}{r}{Conc.} & \multicolumn{2}{r}{Repet.} \\
  &    & Method &  &  &  &  &  &  &  &  &  &  &  &  &  &  &  &  \\
\midrule
\multirow[l]{6}{*}{\rotatebox{90}{Extractive}} & \multirow[l]{4}{*}{ } & Random & 0.32 & \color[HTML]{A0A1A3} ±0.05 & 0.06 & \color[HTML]{A0A1A3} ±0.02 & 0.15 & \color[HTML]{A0A1A3} ±0.03 & 0.09 & \color[HTML]{A0A1A3} ±0.04 & 0.48 & \color[HTML]{A0A1A3} ±0.07 & 0.58 & \color[HTML]{A0A1A3} ±0.12 & 0.15 & \color[HTML]{A0A1A3} ±0.04 & 0.94 & \color[HTML]{A0A1A3} ±0.04 \\
 &  & LSA & 0.38 & \color[HTML]{A0A1A3} ±0.06 & 0.08 & \color[HTML]{A0A1A3} ±0.03 & 0.16 & \color[HTML]{A0A1A3} ±0.03 & 0.14 & \color[HTML]{A0A1A3} ±0.09 & 0.57 & \color[HTML]{A0A1A3} ±0.09 & 0.59 & \color[HTML]{A0A1A3} ±0.15 & 0.19 & \color[HTML]{A0A1A3} ±0.07 & 0.94 & \color[HTML]{A0A1A3} ±0.07 \\
 &  & LexRank & 0.38 & \color[HTML]{A0A1A3} ±0.06 & \bfseries 0.09 & \color[HTML]{A0A1A3} ±0.03 & 0.17 & \color[HTML]{A0A1A3} ±0.03 & 0.18 & \color[HTML]{A0A1A3} ±0.12 & 0.56 & \color[HTML]{A0A1A3} ±0.09 & \bfseries 0.60 & \color[HTML]{A0A1A3} ±0.16 & 0.21 & \color[HTML]{A0A1A3} ±0.08 & \bfseries 0.95 & \color[HTML]{A0A1A3} ±0.06 \\
 &  & QT & 0.21 & \color[HTML]{A0A1A3} ±0.06 & 0.04 & \color[HTML]{A0A1A3} ±0.02 & 0.12 & \color[HTML]{A0A1A3} ±0.03 & 0.09 & \color[HTML]{A0A1A3} ±0.08 & 0.35 & \color[HTML]{A0A1A3} ±0.11 & 0.13 & \color[HTML]{A0A1A3} ±0.05 & 0.12 & \color[HTML]{A0A1A3} ±0.07 & 0.80 & \color[HTML]{A0A1A3} ±0.14 \\
\cline{2-19}
 & \multirow[l]{2}{*}{  } & GLIMPSE-Speaker & 0.22 & \color[HTML]{A0A1A3} ±0.06 & 0.04 & \color[HTML]{A0A1A3} ±0.02 & 0.11 & \color[HTML]{A0A1A3} ±0.03 & 0.09 & \color[HTML]{A0A1A3} ±0.05 & 0.36 & \color[HTML]{A0A1A3} ±0.09 & 0.36 & \color[HTML]{A0A1A3} ±0.13 & 0.13 & \color[HTML]{A0A1A3} ±0.05 & 0.89 & \color[HTML]{A0A1A3} ±0.08 \\
 &  & GLIMPSE-Unique & 0.27 & \color[HTML]{A0A1A3} ±0.06 & 0.06 & \color[HTML]{A0A1A3} ±0.03 & 0.13 & \color[HTML]{A0A1A3} ±0.03 & 0.13 & \color[HTML]{A0A1A3} ±0.07 & 0.39 & \color[HTML]{A0A1A3} ±0.09 & 0.38 & \color[HTML]{A0A1A3} ±0.13 & 0.15 & \color[HTML]{A0A1A3} ±0.06 & 0.89 & \color[HTML]{A0A1A3} ±0.07 \\
\cline{1-19} \cline{2-19}
\multirow[l]{5}{*}{\rotatebox{90}{Abstractive}} & \multirow[l]{3}{*}{ } & PlanSum & 0.25 & \color[HTML]{A0A1A3} ±0.08 & 0.06 & \color[HTML]{A0A1A3} ±0.02 & 0.14 & \color[HTML]{A0A1A3} ±0.04 & 0.07 & \color[HTML]{A0A1A3} ±0.04 & 0.21 & \color[HTML]{A0A1A3} ±0.07 & 0.32 & \color[HTML]{A0A1A3} ±0.11 & 0.13 & \color[HTML]{A0A1A3} ±0.07 & 0.37 & \color[HTML]{A0A1A3} ±0.36 \\
 &  & Coop & 0.36 & \color[HTML]{A0A1A3} ±0.05 & 0.08 & \color[HTML]{A0A1A3} ±0.02 & \bfseries 0.19 & \color[HTML]{A0A1A3} ±0.02 & 0.08 & \color[HTML]{A0A1A3} ±0.09 & 0.26 & \color[HTML]{A0A1A3} ±0.12 & 0.51 & \color[HTML]{A0A1A3} ±0.23 & 0.09 & \color[HTML]{A0A1A3} ±0.07 & 0.26 & \color[HTML]{A0A1A3} ±0.31 \\
 &  & Llama 7b Instruct & \bfseries 0.39 & \color[HTML]{A0A1A3} ±0.06 & 0.09 & \color[HTML]{A0A1A3} ±0.03 & 0.18 & \color[HTML]{A0A1A3} ±0.02 & 0.23 & \color[HTML]{A0A1A3} ±0.12 & \bfseries 0.63 & \color[HTML]{A0A1A3} ±0.11 & 0.49 & \color[HTML]{A0A1A3} ±0.16 & 0.25 & \color[HTML]{A0A1A3} ±0.09 & 0.79 & \color[HTML]{A0A1A3} ±0.26 \\
\cline{2-19}
 & \multirow[l]{2}{*}{  } & GLIMPSE-Speaker & 0.33 & \color[HTML]{A0A1A3} ±0.05 & 0.07 & \color[HTML]{A0A1A3} ±0.03 & 0.15 & \color[HTML]{A0A1A3} ±0.02 & 0.32 & \color[HTML]{A0A1A3} ±0.10 & 0.44 & \color[HTML]{A0A1A3} ±0.06 & 0.53 & \color[HTML]{A0A1A3} ±0.10 & 0.27 & \color[HTML]{A0A1A3} ±0.06 & 0.90 & \color[HTML]{A0A1A3} ±0.08 \\
 &  & GLIMPSE-Unique & 0.34 & \color[HTML]{A0A1A3} ±0.04 & 0.07 & \color[HTML]{A0A1A3} ±0.02 & 0.16 & \color[HTML]{A0A1A3} ±0.02 & \bfseries 0.33 & \color[HTML]{A0A1A3} ±0.10 & 0.44 & \color[HTML]{A0A1A3} ±0.07 & 0.54 & \color[HTML]{A0A1A3} ±0.10 & \bfseries 0.27 & \color[HTML]{A0A1A3} ±0.06 & 0.84 & \color[HTML]{A0A1A3} ±0.11 \\
\cline{1-19} \cline{2-19}
\bottomrule
\end{tabular}}
\caption{Comparison to metareview motivations using ROUGE scores and estimated human judgment using the SEAHORSE metrics for all  baselines and our templated summaries compared against each document independently. Cov. stands for Main ideas, Attr. for Attribution, Gram. for Grammar, Compr. for Comprehensible, Conc. for Conciseness, and Repet. for Repetition. The best value in each column is in bold.}
\label{tab:seahorse}
\end{table*}

\noindent\textbf{Discriminativenes versus fluency.} In \autoref{fig:trade_off_info_fluency}, we plot the discriminativeness (in terms of discriminativenes) against the language model perplexity (as a proxy to fluency). We highlight the Pareto frontier, \textit{i.e.} the points for which we cannot gain on a given axis without losing on the other. Along this frontier, we observe a trade-off between discriminativeness and fluency (\autoref{fig:trade_off_info_fluency}). This outcome is expected, as the RSA scores are designed to select less common utterances to improve discriminativeness.

\begin{table}
    \centering
    \resizebox{\linewidth}{!}{
    \begin{tabular}{l|c|c}
    Method & Discriminativeness & Avg. Summary Length\\
    \midrule
    {\sys} & 93.75\% & 111 \\
    Llama & 85.18\% & 1920 \\
    MDS & 0\% & 268 \\
    \end{tabular}}
    \caption{Accuracy of annotators guessing the review from which a summary is generated for our method and two baselines and the average of summary lengths generated by each method.}
    \label{tab:human_eval_results}
\end{table}

\noindent\textbf{Human evaluation of uniqueness.} Since the main goal of the D-MDS task is to highlight the most unique information from each review, we asked human evaluators to identify the source review for each summary (See~\autoref{sec:human_eval} for details about the human evaluation task). We show that {\sys} selects more informative and unique ideas than other summarization methods such as LLMs (Llama 7b Instruct) or abstractive multi-document summarizers as illustrated 
 in \autoref{tab:human_eval_results}. These results are consistent with the evaluation reported in \autoref{fig:baselines_informativeness}.
They indicate that utterances selected with \sys-Unique are shorter and more discriminative compared with those selected with baseline methods, suggesting its potential as an effective highlighting mechanism for peer review.

\subsection{Overall Summary Quality}
\label{sec:sum_quality}
Using ROUGE scores, we compare the overlap between the generated summaries and the summary-like metareviews. We also evaluate them using the SEAHORSE metrics. We present the results in \autoref{tab:seahorse}.

\noindent\textbf{Comparison to gold standards.} Overall, the summaries generated using both our methods and the baselines exhibit minimal overlap with the meta-reviews. We find that our method achieves comparable performance with baseline methods in terms of ROUGE score. This result is likely influenced by the nature of metareviews, which extends beyond synthesizing reviews to justify a paper's acceptance decision.

\noindent\textbf{Coverage and conciseness.} We evaluate the coverage and conciseness of our generated summaries with the baseline methods. We measure the coverage by assessing if the main ideas of a review are present in the summary using the SEAHORSE metrics (cf. \autoref{tab:seahorse}). 
We observe that the summaries generated with \sys are more concise and yield a significantly better coverage of the main ideas of the documents than baseline methods. This result is consistent with the discriminativeness results presented in (\autoref{fig:baselines_informativeness_per_char} and \autoref{sec:discriminative_summ}). Using {\sys} to select excerpts leads to denser summaries than baseline methods.

\noindent\textbf{Fluency.} Similarly to our findings in the discriminative summarization setting, we find that summaries generated by our method tend to be less fluent than those generated by the baselines. We posit that this is due to our method's tendency to select rarer utterances, aiming to enhance discriminativeness, and thus follows the same trade-off observed in \autoref{fig:trade_off_info_fluency}.

\section{Summary and Concluding Remarks}

We have introduced a discriminative summarization framework for multi-document summarization and suggested a pragmatic summarization approach inspired by the Rational Speech Act model (RSA) of human communication. Our findings demonstrate the effectiveness of RSA-based scores in capturing unique, common, and divergent opinions among reviewers. This paves the way for the development of tools to aid area chairs in evaluating reviews. Our method yields more informative summaries compared to existing approaches and is suitable for extractive summarization, ensuring the interpretability of the generated summaries.

\section*{Acknowledgments}

This work was granted access to the HPC resources of IDRIS under the allocation 2023-AD011013290R2 made by GENCI. Ines Arous is supported by a grant from the former Twitter, Inc. In addition, we acknowledge material support from NVIDIA Corporation in the form of computational resources provided to Mila.

\newpage

\section{Limitations \& Ethical Considerations}

We adopt summarization evaluation methods common in the research area, but the validity of these methods needs further investigation. For example, measuring the overlap between the templated summaries and the metareview motivations is limited by the appropriateness of the motivation as a gold standard and by the use of ROUGE scores, which have well-known limitations. The SEAHORSE trained metrics are also working in a somewhat out-of-distribution setting as they have not been specifically validated on scientific content. Since our method's main strength is to highlight unique and common ideas in a given text to help the reader get the most salient points in context, a natural follow-up work will be to conduct a task-based human evaluation of the highlighting it provides.

Still, extractive summarization methods are notably sensitive to the sentence segmentation process, which can occasionally result in peculiar outcomes. For instance, a brief sentence containing nonsensical content might erroneously emerge as the most informative segment of the summary simply because it is unique among the reviews. Similarly, filler sentences such as "Here are some comments" could be mistakenly identified as the most common ideas in the corpus. Although extractive summarization ensures attribution, it is constrained by these factors, unlike abstractive summarization methods, which, while prone to potential hallucinations, circumvent such pitfalls.

As with any automated tool, its deployment carries inherent risks, particularly if used without due caution in critical decision-making processes, such as determining the acceptance or rejection of scholarly work. Indeed, unexpected biases could lead to discrimination and unfair presentation of the reviewers' points. It is important to note that our approach is not designed to fully automate decision-making or replace the review summarization process altogether. Rather, we advocate for its use as a supplementary tool, particularly in its extractive configuration, to aid in identifying salient points within reviews.

\newpage
\bibliography{zotero}

\appendix

\newpage

\section{Human evaluation}
\label{sec:human_eval}
We recruited expert human annotators in academia to evaluate the informativeness of the produced summaries (already enrolled PhD and Master students in computer science and machine learning). They were compensated based on their skills and location (Montréal, Quebec) at a rate of 30 CAD per hour.

\subsection{Task description}

\begin{figure*}
    \centering
    \includegraphics[width=0.9\linewidth]{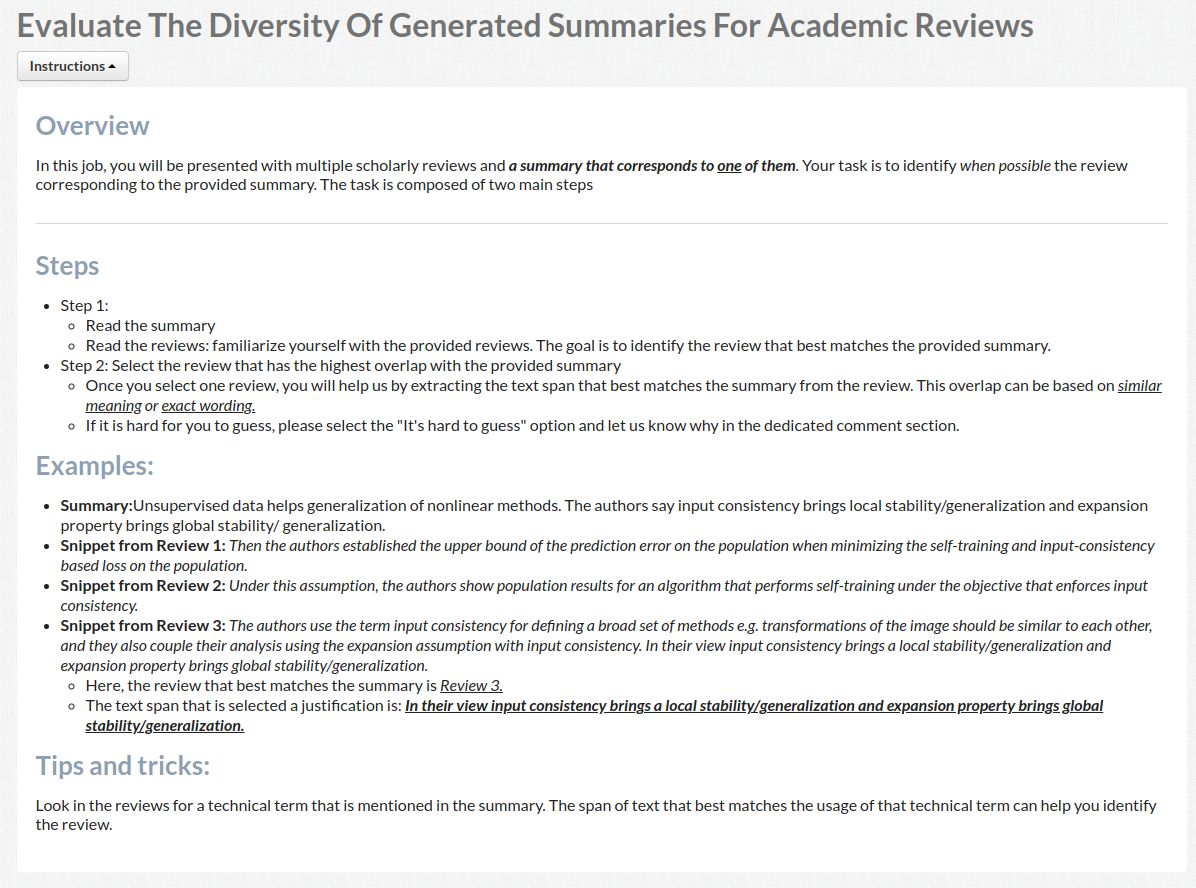}
    \caption{Instructions given to the annotators to perform the discriminative summarization task.}
    \label{fig:human_eval_instructions}
\end{figure*}

\begin{figure*}
    \centering
    \includegraphics[width=0.9\linewidth]{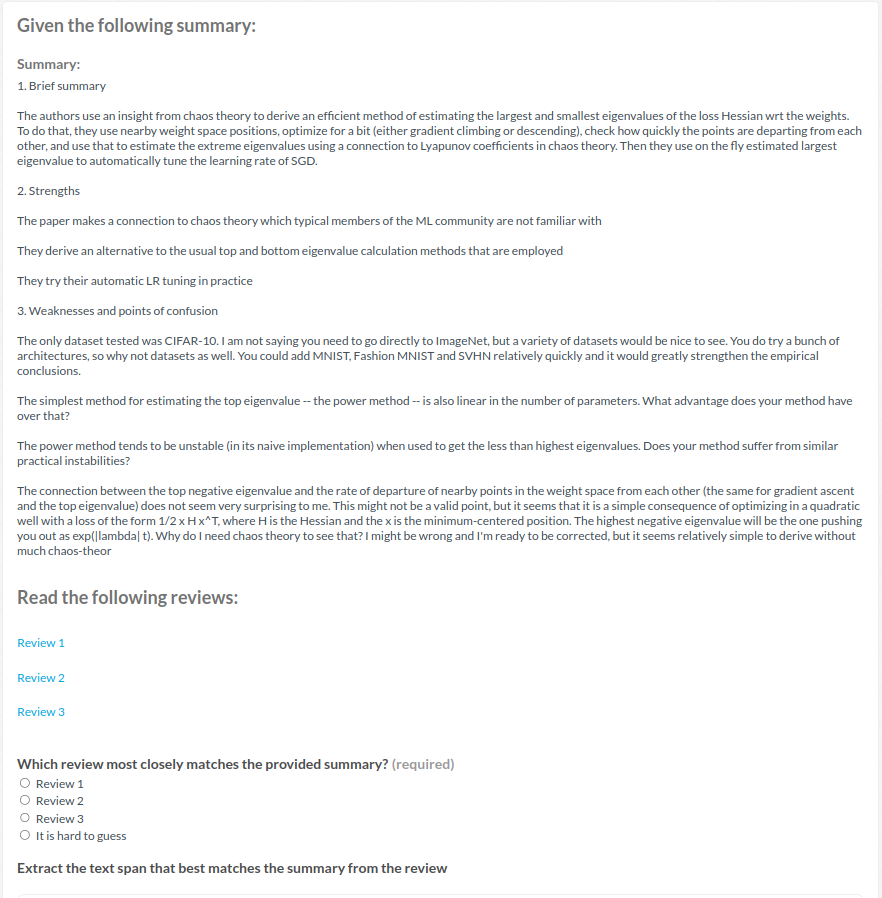}
    \caption{Example of task}
    \label{fig:human_eval_example}
\end{figure*}

The annotators were presented a generated summary and they had to guess which review was input to the summarizer to produce said summary. They were given 4 choices: one per review and the option to say that it's hard to guess. In \autoref{fig:human_eval_instructions} we present the instructions given to the annotators and an example of the actual task in \autoref{fig:human_eval_example}.

\subsection{Human evaluation statistics}

\begin{figure*}
\begin{subfigure}{0.3\linewidth}
    \centering
    \includegraphics[width=\linewidth]{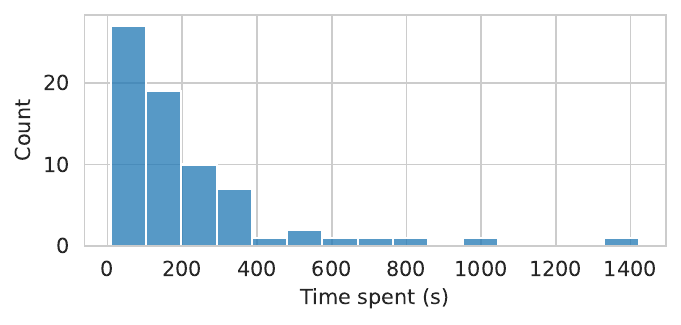}
    \caption{Distribution of the time spent by the annotators.}
    \label{fig:human_eval_time_spent}
\end{subfigure}
\hfill
\begin{subfigure}{0.3\linewidth}
    \centering
    \includegraphics[width=\linewidth]{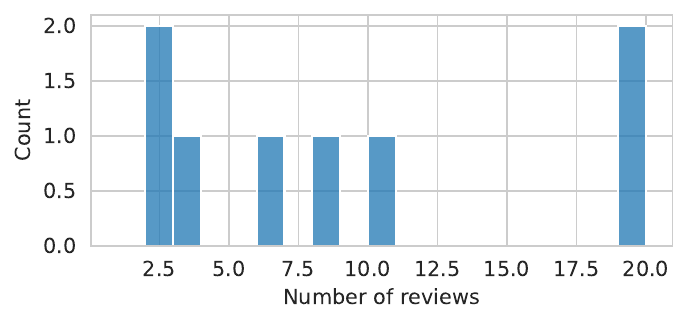}
    \caption{Number of samples examined by each worker.}
    \label{fig:human_eval_number_of_review_worker}
\end{subfigure}
\hfill
\begin{subfigure}{0.3\linewidth}
    \centering
    \includegraphics[width=\linewidth]{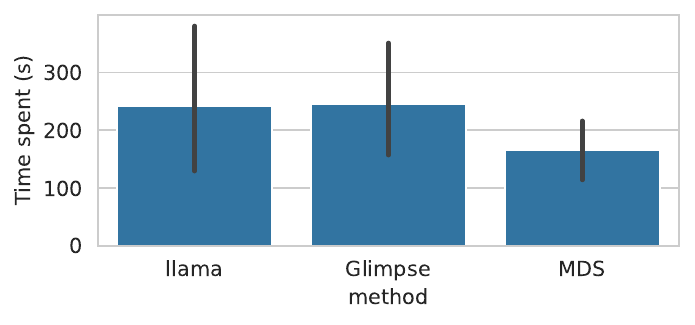}
    \caption{Time spent by the reviewer per method evaluated.}
    \label{fig:human_eval_time_per_method}
\end{subfigure}
\caption{Statistics of the annotators involved.}
\label{fig:human_eval_stats}
\end{figure*}

We got $8$ annotators to perform a total of $89$ evaluations total. They spent on average less than an hour on the task. With some dedicated annotators spending almost two hours and performing up to $20$ evaluations (\autoref{fig:human_eval_stats}).

\section{Impact of the generation method}

\begin{figure}
    \centering
    \includegraphics[width=0.45\textwidth]{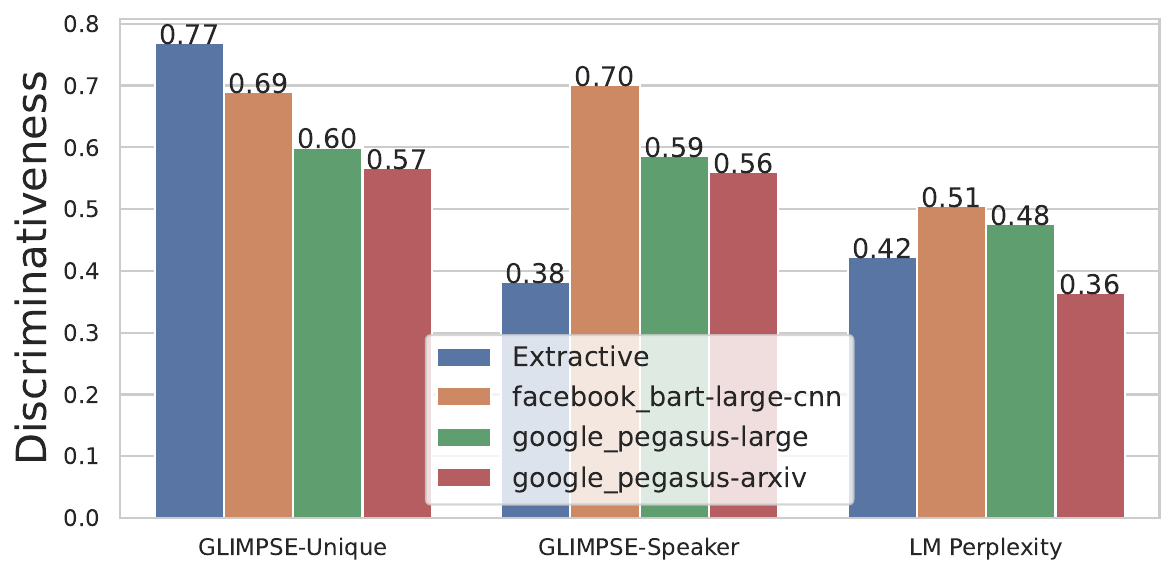}
    \caption{Informativeness of the summaries selected using different scoring methods ({\sys}-Unique, {\sys}-Speaker) and the baseline LM Perplexity from different set of candidates. Candidates generated by different generative models or simply sentences extracted from the document.}
    \label{fig:informativeness}
\end{figure}

 \noindent\textbf{Abstractive summarization.} {\sys} can rely on different generation methods (either extractive or different generative summarizers). In \autoref{fig:informativeness} we compare the informativeness induced by the different generative methods. It seems that BART produces the most promising pool of candidates.

\end{document}